\documentclass[11pt,a4paper]{article}
\usepackage{times,latexsym}
\usepackage{url}
\usepackage[T1]{fontenc}
\usepackage[acceptedWithA]{tacl2021v1}
\usepackage{personal_settings}

\title{\textsc{Hummus}: A Dataset of Humorous Multimodal Metaphor Use}
\author{
Xiaoyu Tong$^\diamond$,
Zhi Zhang$^\diamond$,
Pia Sommerauer$^\dagger$,
Martha Lewis$^\diamond$
\and
Ekaterina Shutova$^\diamond$
\\
\ \\
$^\diamond$ILLC, University of Amsterdam, the Netherlands
\\
$^\dagger$Vrije Universiteit Amsterdam, the Netherlands \\
\texttt\{x.tong,m.a.f.lewis,e.shutova\}@uva.nl, zzhang2626@gmail.com, pia.sommerauer@vu.nl
}
\date{}

\begin{document}
\maketitle
\begin{abstract}
Metaphor and humor share a lot of common ground, 
and metaphor is one of the most common humorous mechanisms.
This study focuses on the humorous capacity of multimodal metaphors, which has not received due attention in the community.
We take inspiration from the Incongruity Theory of humor, the Conceptual Metaphor Theory, and the annotation scheme behind the VU Amsterdam Metaphor Corpus, and developed a novel annotation scheme for humorous multimodal metaphor use in image-caption pairs.
We create the \textsc{Hummus} Dataset of \textbf{Hu}morous \textbf{M}ultimodal \textbf{M}etaphor \textbf{Us}e, providing
expert annotation on 1k image-caption pairs sampled from the New Yorker Caption Contest corpus. 
Using the dataset, we test state-of-the-art multimodal large language models (MLLMs)
on their ability to detect and understand humorous multimodal metaphor use.
Our experiments show that current MLLMs
still struggle with processing humorous multimodal metaphors,
particularly with regard to integrating visual and textual information.
We release our dataset and code at \url{github.com/xiaoyuisrain/humorous-multimodal-metaphor-use}.
\end{abstract}

\section{Introduction}

\begin{figure}[t]
  \includegraphics[width=\columnwidth]{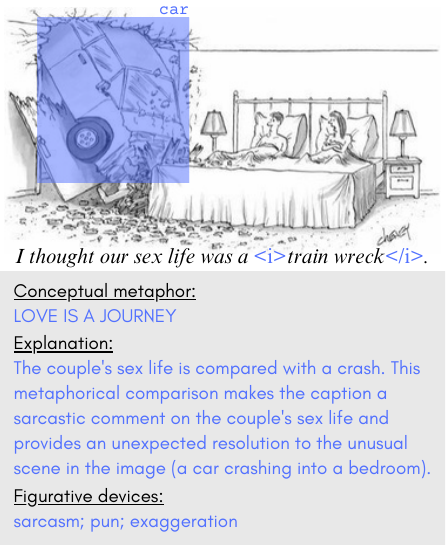}
  \caption{\textsc{Hummus} metaphor sample, with \textcolor{RoyalBlue}{annotation} for image (bounding box and label), caption (\texttt{<i></i>}), conceptual metaphor, explanation of how the metaphor use contributes to humor, and additional figurative devices.}
  \label{fig:train-wreck}
\end{figure}

Conceptual Metaphor Theory (CMT; \citealp{lakoff-johnson-1980-metaphors}) contends that the metaphors we use in language are based on cross-domain mappings in our mind; metaphor is essentially a cognitive process of understanding one thing in terms of another.
For example, when one says \lingform{Our marriage has \underline{gone off the track}} 
, one is using the \textsc{love is a journey} metaphor, conceptualizing \textsc{love} or \textsc{relationship} in terms of a \textsc{journey}.
CMT marks a turning point in metaphor research, shifting focus towards how metaphors are represented or processed in the mind and the brain (see \citealp{holyoak-stamenkovic-2018-metaphor} for a survey).
There has also been an increased interest in nonverbal and multimodal manifestations of metaphor, such as metaphor in films, cartoons, adverts, and gestures \cite{forceville-2015-visual,forceville-2017-visual,kappelhoff-muller-2011-embodied,tsakona-2009-language}.

The influence of this cognitive turn in metaphor research has extended to the field of natural language processing.
The VU Amsterdam Metaphor Corpus (VUA; \citealp{steen-etal-2010-mipvu}), which is created by cognitive linguists and employs an annotation scheme that closely follows the CMT, has inspired research on automatic detection of metaphors in text 
(see \citealp{tong-etal-2021-recent} for a survey).
There is also recent work on computational modelling of visual and multimodal metaphors in adverts, memes, and videos \cite{alnajjar-etal-2022-ring,zhang-etal-2021-multimet,zhang-etal-2023-multicmet,xu-etal-2022-metmeme}.

Metaphor use serves many purposes in communication, including the delivery of humor \cite{michelli-etal-2024-framework,attardo-2015-humorous}.
Metaphor and humor share a lot of common ground.
The Incongruity Theory explains that humor arises from the perception of incongruity, something that violates one's expectations \cite{clark-1970-humour,morreall-2024-philosophy}. Similarly, metaphor researchers consider incongruity as a major contextual clue for identifying metaphors in text \cite{cameron-2003-metaphor,steen-etal-2010-mipvu}.
Metaphor and humor have even been suggested to share neural pathways in the brain \cite{hellberg-2018-funny}, and visual metaphor is found to be one of the most common humorous mechanisms in cartoons \cite{tsakona-2009-language}.

This study is concerned with the humorous capacity of multimodal metaphors, and sheds light on how well multimodal large language models (MLLMs) understand humorous multimodal metaphor use.
We created a Dataset of Humorous Multimodal Metaphor Use (\textsc{Hummus}), providing expert annotation on 1k
image-caption pairs sampled from the New Yorker Caption Contest (CapCon) corpus \cite{hessel-etal-2023-androids} in terms of (1) whether the image-caption pair contains humorous multimodal metaphor use, (2) the conceptual metaphors involved, (3) image and text parts related to the metaphor use,
(4) how the metaphor use contributes to humor,
and (5) use of other figurative devices (e.g., idiom, irony, hyperbole).
An example is provided in Figure~\ref{fig:train-wreck}.
Our final dataset contains 580 metaphorical and 365 non-metaphorical items.

Based on the annotations, we created 6 tasks to test MLLMs' ability to identify and understand humorous multimodal metaphor use: Classification, Naming (of conceptual metaphors), ImageBbox, ImageLabel, CaptionHL, and Explanation.
We tested 7 open-source and 2 closed source MLLMs:
Qwen3-VL-32/8/2B-Instruct,
Qwen2-VL-72/7B-Instruct,
LLaVA-NeXT-110/8B,
GPT-4o, and GPT-4 Turbo, finding that the models struggle with distinguishing metaphorical and non-metaphorical items, as well as understanding the humorous multimodal metaphor use involved.
Nevertheless, it is possible to improve their performance through fine-tuning.
Our ablation study and error analysis reveal that the models' struggles are likely to be caused by difficulties in integrating visual and textual information.

\section{Background}

\paragraph{Metaphor.}
The Conceptual Metaphor Theory posits the existence of conceptual metaphors, mappings in the human mind between different conceptual domains \cite{lakoff-johnson-1980-metaphors}.
Metaphor as a figure of speech is manifestation of conceptual metaphors in language use.
Conceptual metaphors also give rise to visual metaphors (e.g., representing an idea as a light bulb) and multimodal metaphors, which involve more than one mode of communication.

The image-caption pair in Figure~\ref{fig:train-wreck} is a multimodal instantiation of the \textsc{love is a journey} metaphor.
The caption explicitly compares sex life, which belongs to the \textsc{love} domain,  with a train wreck, which can happen in a \textsc{journey}.
The image depicts the \textsc{love} domain with a couple in bed, and the \textsc{journey} domain with a car crashing into the couple's bedroom.
The metaphor use involves both the linguistic and the visual mode, and thus qualifies as multimodal metaphor use.

\paragraph{Figurative language.}
Metaphor, simile, personification, and zoomorphism are different types of figurative language, but they are all considered as metaphors under CMT.
Similes are linguistic metaphors that use words such as \lingform{like} and \lingform{as} to highlight a cross-domain mapping (e.g., \lingform{my love's like a red, red rose});
metaphor as a figure of speech uses categorization statements, making the cross-domain mapping more implicit
(e.g., \lingform{my love is a red, red rose}).
Personification is metaphor that compares something non-human to humans (e.g., pets as family members).
Zoomorphism is metaphor with \textsc{animals} or a species of animal as the source domain (e.g., \lingform{the \underline{roar} of the ocean}).

Idioms are sometimes called dead metaphors, expressions that have lost their metaphoricity over time.
There is empirical evidence that people process idioms and metaphors differently \cite{gibbs-1992-what,desai-2022-metaphors}.
We therefore regard simile, personification, and zoomorphism as metaphor use, but treat idiom as a different phenomenon.

\paragraph{Humor.}
The Incongruity Theory of Humor can be traced back to \citet{beattie-1779-essays} and is the most influential theory for analyzing and explaining humor.
The theory contends that laughter arises from the presence of incongruity, something that \enquote{violates our standard mental patterns and normal expectations \cite{morreall-2024-philosophy}}.
\citet{suls-1972-two} proposes a two-stage model for humor appreciation: perception of incongruity and resolution of incongruity.
For example, the cartoon image in Figure~\ref{fig:train-wreck} is incongruous as car crash normally does not happen in a bedroom.
The caption resolves the incongruity by attributing the car crash to the couple's disastrous relationship, and humor arises from this witty, unexpected attribution.

\section{Related Works}

\paragraph{Linguistic metaphor.}
A large proportion of research on computational modelling of metaphor is restricted to linguistic metaphors, and automatic metaphor detection in particular \cite{li-etal-2023-framebert,maudslay-teufel-2022-metaphorical,shutova-2015-design,tong-etal-2021-recent,wang-etal-2023-metaphor,zhang-liu-2022-metaphor}.
Most metaphor detection studies employ the VU Amsterdam Metaphor Corpus (VUA; \citealp{steen-etal-2010-mipvu}), a four-million word corpus where every word is annotated in terms of whether its use is metaphorical.

Nevertheless, recent years have seen an increased interest in metaphor understanding.
More paraphrase datasets are released \cite{joseph-etal-2023-newsmet,tong-etal-2024-metaphor}, following earlier works that treat metaphor understanding as a paraphrasing task \cite{bizzoni-lappin-2018-predicting,shutova-2010-automatic}.
Other studies frame the issue as an inference task, providing entailed and non-entailed \cite{comsa-etal-2022-miqa,stowe-etal-2022-impli} or contradictory statements \cite{chakrabarty-etal-2022-flute,liu-etal-2022-testing} for metaphorical sentences.
A framework for
intentions behind metaphor use has also been proposed \cite{michelli-etal-2024-framework}.

\paragraph{Visual and multimodal metaphor.}
Automatic identification of text-based metaphors in videos \cite{alnajjar-etal-2022-ring} and figurative language use in memes \cite{liu-etal-2022-figmemes} has been tackled.
For metaphor understanding, the V-FLUTE dataset \cite{saakyan-etal-2024-vflute} provides automatically generated explanations for visual metaphors. 
Other datasets provide detailed annotations for monomodal and multimodal metaphors in social media and adverts \cite{zhang-etal-2021-multimet,zhang-etal-2023-multicmet} or memes \cite{xu-etal-2022-metmeme}.
Their annotations concern metaphor occurrence, modality (whether the metaphor is text-based, image-based, or multimodal), target domain, source domain, intent, and sentiment category.

\citet{akula-etal-2023-metaclue} annotate
visual metaphors in adverts and introduce MetaCLUE, a set of five visual metaphor processing tasks:
classification (whether a given image contains metaphor), localization (identifying image regions), understanding (interpreting metaphors in the form \enquote{<target> is as <property> as <source>}), and generation (generating visual metaphors from text prompt).

\paragraph{Humor and metaphor.}
\citet{hessel-etal-2023-androids} propose a set of humor understanding tasks based on their New Yorker Caption Contest (CapCon) corpus.
The corpus contains earlier releases \cite{jain-etal-2020-new,radev-etal-2016-humor,shahaf-etal-2015-inside} of \textit{The New Yorker} cartoons and captions.
It also provides manually created explanations for 651 image-caption pairs.

\citet{chang-etal-2024-nykms} annotate a subset of the CapCon corpus and propose NYK-MS, a benchmark for metaphor and sarcasm understanding.
While our work also uses the CapCon corpus, our contributions are different in the following ways:
(1) Every metaphorical item in \textsc{Hummus} qualifies as multimodal metaphor use, and its multimodality is reflected in the annotations; NYK-MS, on the other hand, is primarily concerned with metaphorical words in captions.
(2) We annotate the conceptual metaphors underlying the identified multimodal metaphor use; NYK-MS lacks this type of annotation.
(3) We provide bounding box annotations for image areas related to the identified metaphor use, which is also absent in NYK-MS.
(4) We explain our metaphorical items in relation to the humorous effect of the image-caption pairs, whereas NYK-MS does not consider the interplay between metaphor and humor.
(5) \textsc{Hummus} is annotated by a linguist specializing in metaphor research while the ground truth of NYK-MS is based on GPT-4V generations.

\section{Dataset Creation}

\textsc{Hummus} is built upon the image-caption pairs in the CapCon corpus \cite{hessel-etal-2023-androids}. 
Each week, \textit{The New Yorker} publishes a captionless cartoon and receives caption submissions from readers.
The CapCon corpus includes 2578 funny caption submissions for 679
cartoons of \textit{The New Yorker}.
We randomly sampled 251 cartoons from the corpus, for which a total of 1000 image-caption pairs are available. Each image-caption pair is a unique instance in terms of its humor and possible metaphor use.
We designed an annotation scheme informed by theories of humor understanding and metaphor use, and
manually annotated the image-caption pairs on Labelbox.
Our annotation scheme can be summarized into two stages:
(1) humorous metaphor identification, which takes less than a minute per item, and (2) detailed metaphor annotation, which takes $\sim$15 minutes per item. 

\subsection{Humorous metaphor identification}

Humorous metaphor use is identified in two steps: humor understanding and metaphor identification.
Image-caption pairs are tagged \enquote{Yes}, \enquote{No}, \enquote{WIDLII (While In Doubt, Leave It In)}, or \enquote{Discard} at this stage.
Given an image-caption pair, the annotator first employs the incongruity-resolution approach to understand the humor:
They note down all possible incongruities in the image and see how the caption resolves those incongruities.
If the annotator fails to understand the humor (incongruities remain unresolved), the item is tagged as \enquote{Discard}.

If the annotator understands the humor, they proceed to determine whether the humor involves metaphor use---whether it can be attributed to any cross-domain mapping, or a process of conceptualizing/depicting one thing in terms of another.
If the annotator is certain that metaphor use is involved---in other words, they already have an idea of which conceptual metaphor is used---they give a \enquote{Yes} label to the image-caption pair.
On the other hand, the annotator may either
\textit{a)} feel certain that the humor has nothing to do with underlying conceptual metaphors, or
\textit{b)} suspect that the humor involves underlying conceptual metaphors, but cannot immediately specify the target and source domains.
In such cases, the item is labeled as \textit{a)} \enquote{No} and \textit{b)} \enquote{WIDLII} respectively.
Both \enquote{Yes} and \enquote{WIDLII} items participate in the subsequent stage of detailed metaphor annotation.

We adopt this strategy from the annotation scheme behind the VUA corpus \cite{steen-etal-2010-mipvu} to avoid creating false negatives.
It also helps with efficiency and consistency:
WIDLII items essentially correspond to knowledge gaps in the annotator despite their expertise in metaphor research, and are likely to cluster around specific conceptual metaphors.
Dealing with them after forming a general impression of all the cartoons not only allows for a more efficient workflow, but also ensures that the annotation is consistent across instantiations of the same metaphor.

\subsection{Detailed metaphor annotation}

\begin{figure}[t]
  \includegraphics[width=\columnwidth]{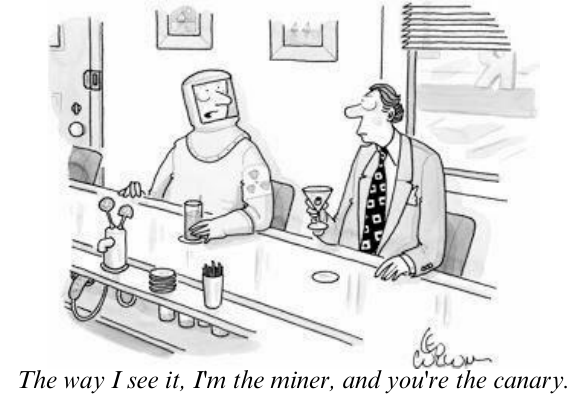}
  \caption{Metaphor sample for which multiple conceptual metaphors are annotated: \textsc{pub is a coal mine}; \textsc{humans are animals}.}
  \label{fig:eg-many-cms}
\end{figure}

Image-caption pairs that are tagged as \enquote{Yes} or \enquote{WIDLII} in terms of the involvement of humorous metaphor use are further annotated in three aspects:
the involved conceptual metaphor, how the conceptual metaphor is reflected in the image (highlighting the relevant objects) and caption (highlighting relevant words), and an explanation of how the metaphor use contributes to the humorous effect of the image-caption pair.
If the annotator can provide full metaphor annotation for a WIDLII item,
it entails that the item indeed involves humorous multimodal metaphor use;
it therefore has the same value as a Yes item.
Otherwise, WIDLII items are re-labeled as \enquote{No} if it becomes clear that no metaphor use is involved; or as \enquote{Discard} if the annotator cannot provide full metaphor annotation, but is still uncertain whether the item qualifies as non-metaphorical.
An explanation is provided if a WIDLII item is eventually Discarded. 

\paragraph{Conceptual metaphor.}
Following the tradition of previous research \cite{lakoff-johnson-1980-conceptual,lakoff-etal-1991-master}, the underlying conceptual metaphors are annotated in \textsc{target domain is source domain} format.

An image-caption pair could employ multiple conceptual metaphors to achieve humorous effects.
The annotator is thus asked to be as inclusive as possible, specifying all conceptual metaphors they can identify.
Take the image-caption pair in Figure~\ref{fig:eg-many-cms} as an example. The image shows two people having drinks in a pub. One of them wears a hazmat suit, which is unheard-of in pubs.
The caption justifies this strange clothing choice (thus resolving the incongruity) by revealing that the person considers himself in a coal mine.
There is thus a metaphorical comparison between \textsc{pub} and \textsc{coal mine}.
The caption also explicitly compares the man in an ordinary suit to a canary in a coal mine (in its literal sense), an instantiation of the \textsc{humans are animals} metaphor.

\paragraph{Image and caption annotation.}
The annotator also marks out image and text fragments that are related to the annotated conceptual metaphor(s).
Metaphor-related image areas are annotated using both bounding boxes and texts (a word or phrase that tells which part of the image is highlighted); each bounding box has a corresponding textual description.
Metaphor-related words or phrases are surrounded by \texttt{<i></i>} tags.

A conceptual metaphor is a mapping between two conceptual domains.
To annotate the image and the caption thus requires the annotator to determine how the two domains are reflected in the two modalities.
Usually, representations of the two domains are unbalanced:
The image should have a recognizable setting that represents the target or the source domain of the metaphor, while a small part of the image (e.g., a particular object) points to the other domain, thus creating incongruity and cross-domain mapping.
Similarly, the context of the caption can be assigned to the target or the source domain, while a particular word or phrase suggests the involvement of the other domain.
The annotator's job is therefore to mark out the less represented domain in the image and the caption respectively.

Let us return to the image-caption pair in Figure~\ref{fig:train-wreck}, which illustrates the \textsc{love is a journey} metaphor.
The image predominantly belongs to the \textsc{love} domain, with a bedroom as the setting and the couple sitting in bed occupying more than half of the image.
The car crashing into the room on the left side of the image evokes the \textsc{journey} domain, creating incongruity and encouraging cross-domain mapping.
While the cracks in the ceiling, the damaged door and the mess on the ground come along with the car crash, they are visual representations of the result of the cross-domain mapping, as opposed to belonging to the source domain, \textsc{journey}, itself.
The car is thus annotated as the metaphor-related fragment in the image.
The caption talks about the couple's sex life and compares it explicitly with a train wreck.
The phrase \lingform{train wreck} is the \enquote{incongruous} part of the caption, evidence of the \textsc{journey} domain in the context of the \textsc{love} domain.
It is therefore marked out as metaphor-related.

\paragraph{Explanation.}
For each image-caption pair that contains humorous multimodal metaphor use, we also provide a short explanation about how the metaphor use contributes to the humor.
This is different from the explanations provided in the CapCon corpus:
While the CapCon corpus focuses on humor understanding and only explicitly mentions metaphor use for two image-caption pairs, our explanations, as exemplified in Figure~\ref{fig:train-wreck}, focus specifically on the interplay between humor and metaphor use. 

\paragraph{Other figurative devices.}
We also provide annotation of the use of other figurative devices, such as pun, idiom, irony, hyperbole, for both metaphorical and non-metaphorical image-caption pairs.
This part of the annotation provides insight into
the relation between humor and the use figurative expressions in general, as well as
co-occurrence of metaphor and other figures of speech in delivering humor. 
We do not use these data to test MLLMs in this study, but still include them in the dataset, as they could be of interest for future research on figurative language.

\subsection{Inter-annotator agreement}

A second expert annotated 300 image-caption pairs randomly sampled from the 1000.
Like the first annotator, the second annotator is a linguist with knowledge of CMT and experience in linguistic metaphor annotation.

To ensure timely completion of the task,
we left out explanation and figurative device annotation and asked the second expert to
(1) classify the items in terms of humorous metaphor use (Yes/No/WIDLII/Discard), (2) identify the conceptual metaphors used in the Yes/WIDLII items, and (3) annotate their image (bounding box) and caption.
To facilitate comparison with model performance, we use the same metrics for inter-annotator agreement calculation and model evaluation: multicalss F1 score for humorous metaphor identification, cosine similarity for conceptual metaphor identification, Intersection over Union (IoU) for image annotation, and Jaccard index for caption annotation.

The co-annotation process started with us introducing the second annotator to our annotation scheme.
We provided them with the written annotation guidelines and verbally walked them through every step of the scheme (excluding explanation and figurative device annotation), exemplifying both metaphorical and non-metaphorical instances.
After the first meeting, the second annotator was assigned the first 50 items to complete independently.
Upon finishing the first 50 items, the two annotators inspected their disagreements together and discussed necessary amendments to the original annotation guidelines.
After the second meeting, we updated the guidelines 
and let the second annotator finish the remaining 250 items independently.
The annotation was completed in 2 months.
With sufficient familiarity with the procedure, annotation for a metaphorical item can be completed in 7 minutes.
The guidelines are provided in Appendix~\ref{sec:annotation-guidelines}.

The final agreement scores were calculated without further discussion of disagreements:
0.73 average F1 score for humorous metaphor identification, with 0.75 for metaphorical (Yes/WIDLII) items and 0.71 for non-metaphorical (No) items;
0.63 mean similarity score for conceptual metaphor identification ($SD=0.20$);
0.73 mean IoU score for image annotation ($SD=0.32$);
0.65 mean Jaccard index for caption annotation ($SD=0.41$).


\begin{figure}[t]
  \includegraphics[width=\columnwidth]{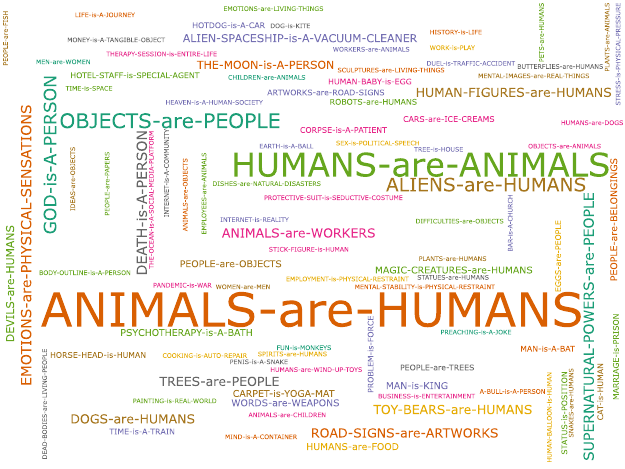}
  \caption{The first 100 conceptual metaphors in our dataset in terms of frequency of occurrence. More frequent ones have larger font sizes.}
  \label{fig:cm-cloud}
\end{figure}

\begin{figure*}[t]
\centering
\begin{subfigure}{0.32\textwidth}
\centering
     \includegraphics[width=0.8\textwidth, trim=0cm 0cm 0cm 0.3cm, clip]{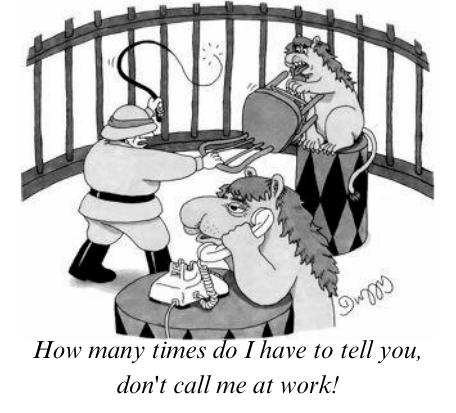}
    \caption{\textsc{animals are humans}.}
    \label{fig:unidirection}
\end{subfigure}
\hfill
\begin{subfigure}{0.32\textwidth}
\centering
    \includegraphics[width=\columnwidth]{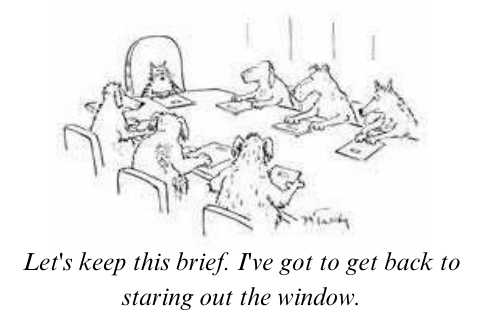}
    \caption{\textsc{animals/humans are humans/animals}.}
    \label{fig:bidirection}
\end{subfigure}
\hfill
\begin{subfigure}{0.32\textwidth}
\centering
    \includegraphics[width=0.8\textwidth, trim=0cm 0cm 0cm 1cm, clip]{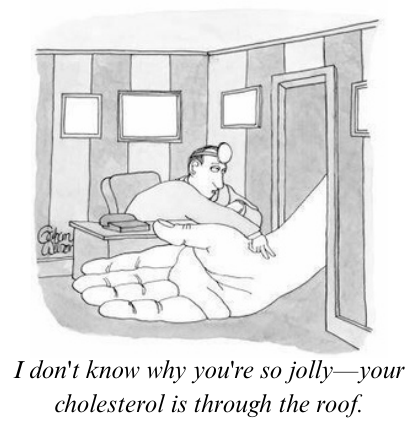}
    \caption{Go through the roof.}
    \label{fig:idiom}
\end{subfigure}
\caption{Example of (a) unidirectional and (b) bidirectional metaphorical mappings between \textsc{animals} and \textsc{humans}, and example of (c) metaphorically used idiom (\lingform{go through the roof}) in our dataset.}
\label{fig:analyzed-samples}
\end{figure*}

\section{Data Analysis}

\textsc{Hummus} provides annotations for 1,000 image-caption pairs, including 580 items that contain humorous multimodal metaphor use (335 \enquote{Yes} and 247 \enquote{WIDLII}\footnote{These WIDLII items are the ones that are makred as \enquote{WIDLII} in the first stage of the annotation procedure (Section 4.1) and receive full metaphor annotation in the second stage (Section 4.2)---and thus not re-labeled as \enquote{No} or \enquote{Discard}. They are just as metaphorical as the Yes items.}), 365 \enquote{No} items, and 55 items marked as \enquote{Discard}.

\paragraph{Multimodality of the metaphor samples.}
Are the metaphors annotated in our dataset identifiable from the captions alone?
To answer this question,
we employ FrameBERT, an automatic textual metaphor detector proposed by \cite{li-etal-2023-framebert}.
The model reaches an F1 score of 0.73 when tested on the VUA-20 dataset \cite{leong-etal-2020-report}.
We consider it as a \enquote{Yes} answer if the model predicts that a given caption contains a content-word real metaphor or borderline metaphor.
Compared with ground truth, the model reaches an accuracy of 0.53 and a F1 score of 0.54, both of which are close to a random guess, and also lower than the model's performance on detecting linguistic metaphors.
This can be considered as an indicator of the multimodal nature of the metaphors annotated in our dataset.

\paragraph{Conceptual metaphor.}
Around 400 conceptual metaphors are identified in the dataset.
As shown in Figure~\ref{fig:cm-cloud},
The most frequently occurring ones are \textsc{animals are humans} (23\% metaphor samples) and \textsc{humans are animals} (8\%).
In addition to 
comparison with \textsc{animals},
the dataset also includes mappings from \textsc{humans} as a source domain to other target domains such as \textsc{object}, \textsc{god}, and \textsc{aliens}.
Personification thus appears to be the most frequently used type of metaphor in these humorous image-caption pairs (51\%).

\begin{table*}[t]
\centering
\small
\renewcommand{\arraystretch}{1.5}
\resizebox{1\textwidth}{!}{
\begin{tabular}{ m{1.9cm} | m{11.1cm} | m{3.2cm} }
\hline
\multicolumn{1}{c|}{\textbf{Task}} & \multicolumn{1}{c|}{\textbf{Instructions}} & \multicolumn{1}{c}{\textbf{Evaluation}} \\
\hline
Classification & Does the humor of the given image-and-caption combination involve metaphor use? Answer the question with Yes or No. & Multiclass F1 score \\
\hline
Naming & The humor of the given image-and-caption combination involves metaphor use. Which conceptual metaphor is used? Answer the question in "TARGET DOMAIN IS SOURCE DOMAIN" format (e.g., "LOVE IS A JOURNEY"). & Sentence similarity \\
\hline
ImageBbox & The humor [...]
involves metaphor use.
Which object in the image is related to the metaphor? Answer with its label and normalized bounding box coordinates in "label: [top, left, height, width]" format. & IoU, precision, recall \\
\hline
ImageLabel & The humor [...]
involves metaphor use.
Which object in the image is related to the metaphor? Answer the question with a single word. & Sentence similarity \\
\hline
CaptionHL & The humor [...]
involves metaphor use. Which part of the caption is related to the metaphor? Surround it with a pair of \texttt{<i></i>} tag. & Jaccard index, precision, recall \\
\hline
Explanation & How does metaphor use contribute to the humor of the given image-and-caption combination? Explain in no more than 30 words. & ROUGE-1, ROUGE-2 \\
\hline
\end{tabular}}
\caption{Benchmark tasks, prompts, and evaluation metrics.
A full prompt includes an image, a caption, and instructions: \texttt{<image>+f\enquote{Caption:\textvisiblespace\{caption\}\symbol{92}n\symbol{92}n\{instructions\}}}.
The first sentence of Image and Caption tasks are the same as the Naming task.
IoU $=$ 
Intersection over Union.}
\label{tbl:tasks}
\end{table*}

\paragraph{Directionality of metaphors.}
A crucial and inevitable step in identifying any (conceptual) metaphor is determining the direction of the metaphorical mapping---in other words, which of the two conceptual domains at play is the target domain, and which is the source domain.
Metaphorical mappings are unidirectional:
Properties and relations of the source domain are projected onto the target domain, not the other way around.
One cannot reverse the direction of a mapping without creating an entirely different metaphor.
Consider the image-caption pair in Figure~\ref{fig:unidirection}, for example.
It is clear that the joke is based on a metaphor with \textsc{animals} as the target domain and \textsc{humans} as the source domain:
A hippo is given human characteristics---it uses telephones and gets angry when someone calls it at an inconvenient time.

This rule of unidirectionality applies to \enquote{prototypical metaphors of all kinds and occurring in all media} \cite{forceville-2002}
while exceptions also exist \cite{carroll-1994,forceville-1995}.
\textsc{Hummus} also demonstrates some exceptions to the rule, especially with regard to the \textsc{humans} and \textsc{animals} domains.
The image-caption pair in Figure~\ref{fig:bidirection}, for example, shows a cat having a meeting with some other animals in 
a modern conference room.
The cat expresses its wish to keep the meeting brief, so that it could go back to staring out the window.
The fact that the animals are sitting in a conference room and engaged in a meeting can be considered as employing the \textsc{animals are humans} metaphor.
The humor of the image-caption pair thus lies in the cat attaching ample importance to staring out the window despite its position in the company; it also invites the reader to wonder what kind of business this company might be running.

On the other hand, one can also interpret the joke as based on the metaphor \textsc{humans are animals}---people working in a company are represented as animals in the image.
By representing the person in a higher position as a cat, the metaphor satirically emphasizes the absurdity of a common scenario in society: A person has an important position in a company while all they care about is something as unproductive as staring out the window.
For this image-caption pair, therefore, it is difficult to distinguish \textsc{humans} and \textsc{animals} in terms of target and source domains, unless we ask the caption writer directly; but it could also happen that the caption writer intended the metaphor to be bidirectional in the first place.
Nonetheless, the double interpretation adds to the depth of such image-caption pairs, making them particularly interesting cases for both metaphor analysis and model evaluation.

\paragraph{Metaphor-related image and caption parts.}
We combine our conceptual metaphor and
image/caption annotations
to estimate whether the metaphor-related parts in image/caption pertain to the target or the source domain. 
For each image-caption pair, we split each conceptual metaphor annotation (e.g., \textsc{animals are humans}) into a target-domain term (\textsc{animals}) and a source-domain term (\textsc{humans}).
We then calculate the similarity between each bounding box label and each of the target- and source-domain terms.
The conceptual domain term yielding the highest similarity score is considered the domain depicted by the metaphor-related image areas.
We also do the same for each highlighted part in the caption.
For calculating the similarity scores, we adopt LaBSE \citep{feng-etal-2022-language} as the evaluator due to its superior performance in matching equivalent conceptual metaphor annotations (Section~\ref{sec:model-eval}).

We find that 60\% of our metaphor samples have image annotations pertaining to target domain, whereas 55\% have caption annotations pertaining to source domain.
The results are in accord with the observation that metaphors usually employ a more concrete domain as the source domain to conceptualize a more abstract target domain.
It is thus plausible that the target domain stands out in visual mode (image) while the source domain stands out in linguisitc mode (caption).

\paragraph{Co-occurrence with other figurative devices.}
More than half (65\%) of the metaphor samples in \textsc{Hummus} feature metaphor use co-occurring with the use of other figurative devices (excluding personification and zoomorphism, which are considered metaphors in this study).
The most frequently used ones include pun (27\%), exaggeration (10\%), and satire (10\%).

We also find a small percentage of metaphor samples (5\%) where the humorous metaphor use concerns an idiom---recall that idioms are usually considered dead metaphors (Section 2).
Consider the image-caption pair in Figure~\ref{fig:idiom}.
When we look at the caption alone, its use of the idiom \lingform{go through the roof} is non-metaphorical:
One understands it refers to a high cholesterol level without visualizing cholesterol actually going through the roof.
When the caption is combined with the image, however, the metaphorical mapping between \textsc{amount} and \textsc{height} is resurrected, and it is precisely the resurrection of the dead metaphor that brings out humor.

\section{Model Evaluation}
\label{sec:model-eval}

\begin{table*}[t]
\centering
\resizebox{0.75\textwidth}{!}{
\begin{tabular}{lcccccc}
\toprule
\multirow{2}[3]{*}{\textbf{Model}} 
& \multicolumn{3}{c}{\textbf{Classification}}
& \textbf{Naming}
& \multicolumn{2}{c}{\textbf{Explanation}} \\
\cmidrule(lr){2-4} \cmidrule(lr){5-5} \cmidrule(lr){6-7}
& Pos & Neg & Avg & SBERT & ROUGE-1 & ROUGE-2 \\
\midrule
Qwen3-VL-32B
& 0.75 & 0.19 & 0.47 & 0.42 \sd{0.17} & 0.24 \sd{0.07} & 0.04 \sd{0.04} \\
Qwen3-VL-8B
& 0.75 & 0.12 & 0.43 & 0.44 \sd{0.17} & 0.22 \sd{0.07} & 0.03 \sd{0.04} \\
Qwen3-VL-2B
& 0.75 & 0.04 & 0.39
& \colorbox{red!90}{0.45 \sd{0.18}}
& 0.26 \sd{0.08} & 0.04 \sd{0.04} \\
\midrule
Qwen2-VL-72B
& 0.64 & 0.43 & 0.54 & 0.44 \sd{0.18} & 0.25 \sd{0.08} & 0.04 \sd{0.04} \\
Qwen2-VL-7B
& 0.61 & 0.46 & 0.53 & 0.40 \sd{0.18} & 0.26 \sd{0.08} & 0.04 \sd{0.04} \\
\midrule
LLaVA-NeXT-110B
& 0.69 & 0.36 & 0.53 & 0.43 \sd{0.18}
& \textbf{0.27 \sd{0.08}} & \textbf{0.05 \sd{0.04}} \\ 
LLaVA-NeXT-8B
& 0.74 & 0.00 & 0.37 & 0.46 \sd{0.18} & 0.26 \sd{0.08} & 0.04 \sd{0.04} \\
\midrule
GPT-4o
& 0.70 & 0.39 & 0.55 & 0.47 \sd{0.22} & 0.23 \sd{0.08} & 0.03 \sd{0.04} \\
GPT-4 Turbo
& 0.64 & 0.47 & \textbf{0.56}
& \colorbox{red!10}{\textbf{0.49 \sd{0.21}}}
& 0.22 \sd{0.08} & 0.03 \sd{0.04} \\
\midrule
\textit{Random} 
& 0.54 & 0.45 & 0.50 &-- &-- &-- \\
\bottomrule
\end{tabular}}
\caption{Classification, Naming, and Explanation benchmarks.
Standard deviations are in parenthesis.
Best model performance per task is in boldface.
Success rate is 100\% except for GPT-4 Turbo \colorbox{red!10}{(96\%)} and Qwen3-VL-2B \colorbox{red!90}{(5\%)} in the Naming task.}
\label{tbl:benchmark-overall}
\end{table*}

\begin{table*}[t]
\centering
\resizebox{0.99\textwidth}{!}{
\begin{tabular}{l ccc c ccc}
\toprule
\multirow{2}[3]{*}{\textbf{Model}} 
& \multicolumn{3}{c}{\textbf{ImageBbox}}
& \textbf{ImageLabel}
& \multicolumn{3}{c}{\textbf{CaptionHL}} \\
\cmidrule(lr){2-4} \cmidrule(lr){5-5} \cmidrule(lr){6-8}
& P & R & IoU & SBERT & P & R & Jaccard \\
\midrule
Qwen3-VL-32B
& \colorbox{red!10}{0.59 \sd{0.38}} & \colorbox{red!10}{0.73 \sd{0.37}} & \colorbox{red!10}{\textbf{0.51 \sd{0.38}}}
& 0.53 \sd{0.30}
& 0.53 \sd{0.39} & 0.80 \sd{0.36} & 0.45 \sd{0.35} \\
Qwen3-VL-8B
& \colorbox{red!10}{0.60 \sd{0.41}} & \colorbox{red!10}{0.59 \sd{0.41}} & \colorbox{red!10}{0.46 \sd{0.38}} 
& 0.51 \sd{0.31}
& 0.44 \sd{0.37} & 0.88 \sd{0.30} & 0.40 \sd{0.33} \\
Qwen3-VL-2B
& \colorbox{red!10}{0.51 \sd{0.42}} & \colorbox{red!10}{0.57 \sd{0.43}} & \colorbox{red!10}{0.42 \sd{0.39}}
& 0.52 \sd{0.31}
& 0.31 \sd{0.32} & 0.98 \sd{0.12} & 0.30 \sd{0.31} \\
\midrule
Qwen2-VL-72B 
& 0.52 \sd{0.44} & 0.41 \sd{0.35} & 0.30 \sd{0.30} & 0.59 \sd{0.28} & 0.40 \sd{0.37} & 0.89 \sd{0.29} & 0.36 \sd{0.34} \\
Qwen2-VL-7B  
& \colorbox{red!10}{0.56 \sd{0.43}} & \colorbox{red!10}{0.43 \sd{0.35}} & \colorbox{red!10}{0.34 \sd{0.32}}
& 0.57 \sd{0.27}
& \colorbox{red!50}{0.36 \sd{0.35}} & \colorbox{red!50}{0.94 \sd{0.22}} & \colorbox{red!50}{0.34 \sd{0.32}} \\
\midrule
LLaVA-NeXT-110B
& \colorbox{red!10}{0.49 \sd{0.39}} & \colorbox{red!10}{0.64 \sd{0.40}} & \colorbox{red!10}{0.41 \sd{0.35}}
& 0.58 \sd{0.27}
& \colorbox{red!20}{0.42 \sd{0.38}} & \colorbox{red!20}{0.93 \sd{0.24}} & \colorbox{red!20}{0.40 \sd{0.37}} \\
LLaVA-NeXT-8B  
& 0.38 \sd{0.36} & 0.77 \sd{0.36} & 0.34 \sd{0.32}
& 0.58 \sd{0.27}
& \colorbox{red!10}{0.33 \sd{0.33}} & \colorbox{red!10}{0.95 \sd{0.20}} & \colorbox{red!10}{0.31 \sd{0.30}} \\
\midrule
GPT-4o 
& 0.40 \sd{0.34} & 0.43 \sd{0.34} & 0.25 \sd{0.22}
& \textbf{0.63 \sd{0.29}}
& 0.47 \sd{0.36} & 0.85 \sd{0.32} & 0.42 \sd{0.32} \\
GPT-4 Turbo
& \colorbox{red!10}{0.38 \sd{0.37}} & \colorbox{red!10}{0.32 \sd{0.34}} & \colorbox{red!10}{0.19 \sd{0.21}}
& 0.56 \sd{0.27}
& 0.54 \sd{0.39} & 0.80 \sd{0.37} & \textbf{0.47 \sd{0.36}} \\
\bottomrule
\end{tabular}}
\caption{ImageBbox, ImageLabel, and CaptionHL benchmarks in mean (SD) format.
Best model performance per task is in boldface.
Some of the tests result in success rates 
\colorbox{red!10}{$\geq 97\%$}, \colorbox{red!20}{$\sim 90\%$},
and \colorbox{red!50}{$< 50\%$}.}
\label{tbl:benchmark-image-caption}
\end{table*}

We design six tasks for humorous multimodal metaphor processing: Classification, Naming, ImageBbox, ImageLabel, CaptionHL, and Explanation.
Table~\ref{tbl:tasks} provides an overview of the instructions and evaluation metrics for each task.
The Classification task includes all items in the test set---that is, the image-caption pair is tagged \enquote{Yes}, \enquote{No}, or \enquote{WIDLII} in terms of whether or not it involves humorous multimodal metaphor use.
The \enquote{No} items are considered negative cases in the Classification task; the \enquote{Yes} and \enquote{WIDLII} items positive, as our dataset provides full metaphor annotation for both categories.
All other tasks only involve the positive cases.

The Naming task and the ImageLabel task employ LaBSE \cite{feng-etal-2022-language} as the evaluator.
Our choice is based on a pilot Naming test that involves the first 100 items in our test set and two models: GPT-4o and GPT-4 Turbo.
We calculate cosine similarity scores between the model outputs and ground truth using a variety of SBERT models. We choose LaBSE as its predictions align the most closely to human judgement of good and bad answers.

We evaluate nine MLLMs: 
Qwen3-VL-32/8/2B-Instruct \citep{qwen3},
Qwen2-VL-72/7B-Instruct \citep{wang-2024-qwen2vl},
LLaVA-NeXT-110/8B \citep{liu2023improvedllava,liu2023llava,liu2024llavanext},
GPT-4 Omni (GPT-4o, \texttt{gpt-4o-2024-05-13}),
and GPT-4 Turbo (\texttt{gpt-4-turbo-2024-04-09}).
These cover state-of-the-art open and closed source models,
as well as smaller models that require less computing resources,
and thus can be more versatile than their larger counterparts in certain use cases (e.g., for fine-tuning).

\subsection{Benchmark results}

Table~\ref{tbl:benchmark-overall} and \ref{tbl:benchmark-image-caption} show results of testing the nine models on the six tasks presented in Table~\ref{tbl:tasks}.
We also report success rates, which measure whether the models provide 
meaningful answers for evaluation 
(e.g., if a model merely repeats the instructions, it fails to provide a meaningful answer).

The random baseline for the Classification task is calculated by randomly choosing between a \enquote{Yes} and a \enquote{No} answer for each item, and averaging the results of 100 iterations.
For ImageBbox, we follow standard practice in object detection \cite{everingham-etal-2010-pascal,lin-etal-2014-microsoft} and consider IoU scores of 0.5 or higher as indicating sufficient overlap between model prediction and ground truth.
For Naming, ImageLabel, and Explanation, we decide thresholds for good answers by manually examining model answers in various score ranges.
The threshold for Naming and ImageLabel is set at 0.6.
For Explanation, model answers reaching a ROUGE-1 score of 0.35 and a ROUGE-2 score of 0.087 are considered acceptable.
Model outputs of different score ranges in these three tasks are presented in Appendix~\ref{sec:more-prompts}.

\paragraph{How well do the models identify humorous multimodal metaphor use?}
All nine models are prone to classifying the image-caption pairs as metaphorical.
While they achieve F1 scores higher than the random baseline in identifying positive cases, even the highest F1 score for the negative category (0.47 by GPT-4 Turbo) is merely around the random baseline (0.45).

\paragraph{How well do the models identify the underlying conceptual metaphors?}
The models struggle with the task to varying degrees;
the average scores are lower than the 0.6 threshold of acceptable answers by 0.11 to 0.20.
While the best performance is achieved by GPT-4 Turbo (0.49),
larger models are not necessarily better at the task.
LLaVA-NeXT-8B, for example, performs slightly better than the 110B model (0.46 vs 0.43).

\paragraph{How well do the models localize humorous multimodal metaphor use in image and caption?}
The Qwen3-VL models exhibit superior ability to draw bounding boxes for metaphor-related image areas (ImageBbox), with the 32B model reaching an average IoU of 0.51, indicating sufficient overlap with ground truth annotations.
The other models, on the other hand, are better at labeling the image areas (ImageLabel), reaching average scores that are close to or surpass the 0.6 threshold.

When highlighting metaphor-related text fragments (CaptionHL),
a common problem among all models is that they tend to include
a longer text fragment than ground truth (the high recall scores).
This indicates that the models lack the ability to sufficiently differentiate instantiations of target and source domains in the captions.

\paragraph{How well do the model explain humorous multimodal metaphor use?}
The Explanation task appears to be difficult for all models.
Merely 5\% of model predictions reach our threshold of acceptable explanations.

\subsection{Prompt engineering}

\begin{table}[t]
\centering
\small
\renewcommand{\arraystretch}{1.1}
\resizebox{0.48\textwidth}{!}{
\begin{tabular}{lccc}
\toprule
 \textbf{Model}         & \textbf{Positive} & \textbf{Negative} & \textbf{Average} \\
\midrule
GPT-4o    & 0.71 \sd{0.01} & 0.43 \sd{0.02} & 0.57 \sd{0.01} \\
Qwen2-VL-72B & 0.61 \sd{0.06} & 0.46 \sd{0.07} & 0.53 \sd{0.01} \\
Qwen2-VL-7B  & 0.66 \sd{0.06} & 0.36 \sd{0.11} & 0.51 \sd{0.03} \\
\bottomrule
\end{tabular}}
\caption{Mean F1 scores using 3 different prompts for the Classification task.
Qwen2-VL stands for Qwen2-VL-Instruct. Success rate is always 100\%.}
\label{tbl:ismet-prompt-engineer-meanf1}
\end{table}

We experiment with other ways to formulate the Classification task, to see whether the high probability of \enquote{Yes} answers is associated with the prompt we use.
Using the first 100 items in the test set, we run a pilot study that tests a wide range of prompts on
Qwen2-VL-7B-Instruct. 
Instead of asking the model to reply with \enquote{Yes} or \enquote{No}, these prompts require different ways to label the given image-caption pair, such as \enquote{Metaphorical/Non-metaphorical}, \enquote{True/False}, \enquote{A/B}.

We determine the top-3 prompts that result in the highest average F1 scores in the pilot study.
These prompts ask the models to reply with
(1) \enquote{Yes (i.e., metaphor use is involved) or No (i.e., metaphor use is not involved)}, (2) \enquote{No or Yes}, and (3) \enquote{A or B}, respectively.
For the third prompt, the two options correspond to whether or not the given image-caption pair involves humorous metaphor use (i.e., \enquote{involves} or \enquote{does not involve}), and the order of the two options is randomised for each test item.

We test these three prompts on three models:
GPT-4o, and the two Qwen2-VL models.
As shown in Table~\ref{tbl:ismet-prompt-engineer-meanf1},
the two Qwen2-VL models, especially the smaller one, are sensitive to different prompts.
The performance of GPT-4o, on the other hand, remains stable over different prompts.
The experiment thus proves the reliability of our Classification benchmark results.

\begin{table*}[t]
\centering
\resizebox{0.75\textwidth}{!}{
\begin{tabular}{lcccccc}
\toprule
\multirow{2}[3]{*}{\textbf{Model}} 
& \multicolumn{3}{c}{\textbf{Classification}}
& \textbf{Naming} & \textbf{ImageBbox} & \textbf{CaptionHL} \\
\cmidrule(lr){2-4} \cmidrule(lr){5-5} \cmidrule(lr){6-6} \cmidrule(lr){7-7}
& Pos & Neg & Avg & SBERT & IoU & Jaccard \\
\midrule
Qwen3-VL-32B
& 0.74 & 0.00 & 0.37
& \textbf{0.54 \sd{0.19}} & 0.34 \sd{0.32} & 0.43 \sd{0.38} \\
Qwen3-VL-8B
& 0.74 & 0.02 & 0.38
& 0.54 \sd{0.20} & \textbf{0.43 \sd{0.35}} & \colorbox{red!10}{0.40 \sd{0.39}} \\
Qwen3-VL-2B
& 0.74 & 0.00 & 0.37
& 0.44 \sd{0.18} & 0.27 \sd{0.29} & \colorbox{red!40}{0.39 \sd{0.35}} \\
\midrule
Qwen2-VL-72B
& 0.73 & 0.00 & 0.37 & 0.47 \sd{0.19} & \colorbox{red!20}{0.22 \sd{0.25}} & \colorbox{red!10}{0.39 \sd{0.39}} \\
Qwen2-VL-7B 
& 0.73 & 0.00 & 0.36 & 0.51 \sd{0.20} & \colorbox{red!10}{0.28 \sd{0.28}} & \colorbox{red!40}{0.38 \sd{0.36}} \\
\midrule
LLaVA-NeXT-110B
& 0.73 & 0.00 & 0.37 & 0.49 \sd{0.19} & \colorbox{red!20}{0.36 \sd{0.28}} & \colorbox{red!40}{\textbf{0.62 \sd{0.43}}} \\ 
LLaVA-NeXT-8B  
& 0.74 & 0.01 & 0.37 & 0.45 \sd{0.17} & 0.20 \sd{0.21} & \colorbox{red!40}{0.30 \sd{0.31}} \\
\midrule
GPT-4o
& 0.70 & 0.36 & \textbf{0.53}
& \colorbox{red!20}{0.53 \sd{0.23}} & \colorbox{red!30}{0.25 \sd{0.20}} & \colorbox{red!30}{0.50 \sd{0.39}} \\
GPT-4 Turbo
& 0.73 & 0.22 & 0.48 & \colorbox{red!10}{0.50 \sd{0.23}} & \colorbox{red!20}{0.21 \sd{0.20}} & \colorbox{red!10}{0.33 \sd{0.33}} \\
\midrule
\textit{Human} & 0.75 & 0.71 & 0.73 & 0.63 \sd{0.20} & 0.73 \sd{0.32} & 0.65 \sd{0.41} \\
\bottomrule
\end{tabular}}
\caption{Model vs human performance on Classification, Naming, ImageBbox, and CaptionHL tasks;
prompts for the models are similar to the annotation guidelines.
Best model performance per task is in boldface.
Some of the tests result in success rates
\colorbox{red!10}{$\geq 95\%$},
\colorbox{red!20}{$\sim 90\%$},
\colorbox{red!30}{$\sim 80\%$},
\colorbox{red!40}{$\geq 60\%$}.}
\label{tbl:model-human-comparison}
\end{table*}

\subsection{Comparison with human performance}
\label{sec:model-human-compare-expt}

Is the models' compromised performance really due to
a lack of capacity to process humorous multimodal metaphor use, or is it due to the subjectivity of the tasks?
To answer this question, we design a new prompt that is similar to the annotation guidelines we provided to the second expert annotator---the models are instructed to complete Classification, Naming, ImageBbox, and CaptionHL in one go, following the same steps as the expert annotator.
We also ask the models to adopt a persona that is representative of the human experts: a linguist who is familiar with CMT and has experience in manual metaphor annotation.
The prompt is provided in Appendix~\ref{sec:more-prompts}.
It is tested on
Qwen2-VL-7B-Instruct before being used for the full experiment.

As shown in Table~\ref{tbl:model-human-comparison}, even when following similar guidelines, none of the models reach the level of agreement between the two expert annotators.
Model performance is closer to human performance in the Naming task, as compared to the other tasks.
In CaptionHL, LLaVA-NeXT-110B reaches a Jaccard index score (0.62) that is close to human (0.65), but with a low success rate (62\%):
for $\sim$40\% of the instances the model identifies as metaphorical, it fails to provide a caption annotation.
In Classification and ImageBbox, the human expert surpasses the models by a large margin.

When compared with the benchmark results, which are obtained through six separate tasks (Table~\ref{tbl:benchmark-overall} and \ref{tbl:benchmark-image-caption}), it is clear that tackling the different tasks in one go benefits the identification of conceptual metaphors, which is evident in the higher Naming scores across all models.
On the other hand, this new approach introduces further challenge to Classification, ImageBbox, and CaptionHL---the models typically perform better when dealing with these tasks separately, despite the lack of detailed instructions on what metaphor is and how to perform the requested annotations.
The comparison suggests that MLLMs and human experts approach the tasks in intrinsically different ways.
For human experts, the tasks inform each other:
Completing Naming, ImageBbox, and CaptionHL annotations help determine whether the image-caption pair is metaphorical (Classification); Naming also provides the basis for ImageBbox and CaptionHL annotations.
For the MLLMs, however, the inherent complexity of producing answers in various formats seems to overwrite the benefit of having additional information.

\begin{table}[t]
\centering
\begin{tabular}{l l l}
\toprule
& \textbf{Train} & \textbf{Test} \\
\midrule
\textbf{All} & 659 & 286 \\
\textbf{Met} & 402 & 178 \\
\bottomrule
\end{tabular}
\caption{Train/test splits for the fine-tuning experiments.
Classification uses the All sets;
the other tasks (Naming, ImageBbox, ImageLabel, CaptionHL, Explanation)
uses the Met sets (i.e., the metaphor samples in the All sets).}
\label{tbl:train-test-splits}
\end{table}

\begin{table}[t]
\centering
\begin{tabular}{l l l}
\toprule
\textbf{Task} & \textbf{0-shot} & \textbf{Fine-tuned} \\
\midrule
Classification & 0.38 & 0.46 \\
Naming & \colorbox{red!90}{0.38 \sd{0.08}} & 0.67 \sd{0.27} \\
ImageLabel & 0.52 \sd{0.31} & 0.62 \sd{0.31} \\
ImageBbox & \colorbox{red!10}{0.38 \sd{0.38}} & 0.61 \sd{0.38} \\
CaptionHL & 0.30 \sd{0.32} & \colorbox{red!20}{0.59 \sd{0.44}} \\
Explanation & & \\
$\hookrightarrow$ \texttt{ROUGE-1} & 0.25 \sd{0.07} & 0.33 \sd{0.12} \\
$\hookrightarrow$ \texttt{ROUGE-2} & 0.04 \sd{0.04} & 0.11 \sd{0.10} \\
\bottomrule
\end{tabular}
\caption{Qwen3-VL-2B-Instruct zero-shot vs fine-tuned results
on the six benchmark tasks:
average F1 score for Classification;
sentence similarity score for Naming and and ImageLabel;
IoU for ImageBbox; Jaccard index for CaptionHL; ROUGE scores for Explanation.
Less than 100\% success rates:
\colorbox{red!10}{99\%},
\colorbox{red!20}{93\%},
\colorbox{red!90}{4\%}.}
\label{tbl:finetune-results}
\end{table}

\subsection{Fine-tuning experiments}

We conduct fine-tuning experiments to test whether MLLMs can learn from our dataset.
We adopt Low-Rank Adaptation (LoRA; \citealp{hu-etal-2021-lora})
to fine-tune Qwen3-VL-2B-Instruct on each of the six benchmark tasks,
using the 30\% \textsc{Hummus} subset annotated by both human annotators as the test set, and the rest (70\%) as training data.
The number of training and test items for the different tasks is provided in Table~\ref{tbl:train-test-splits}.

We consult the official QwenVL training framework\footnote{\url{http://github.com/QwenLM/Qwen3-VL/tree/main/qwen-vl-finetune}}
for our training setup,
but adjust it to the relatively small size of our training dataset.
Specifically, we apply LoRA adapters to the query, key, value, and output projection matrices with $r = 16$, $\alpha = 32$, and no dropout.
We train for 5 epochs with a batch size of 4,
using a learning rate of $1 \times 10^{-4}$ with a cosine scheduler
and a warmup ratio of 0.05.
The hyperpameters are fixed across all tasks.

As shown in Table~\ref{tbl:finetune-results},
fine-tuning results in improved performance across all tasks as compared to the model's zero-shot performance on the same test items following the same prompts.
Most notably, while the model barely manages the Naming task zero-shot (it repeats the example conceptual metaphor in the prompt most of the time; the conceptual metaphors it proposes are also of poor quality),
it produces reliable answers (average score above the 0.6 threshold) consistently (100\% success rate) after fine-tuning.
Similarly huge leaps in performance are also seen in the ImageBbox and CaptionHL tasks.
While the highest scores the model can reach require further optimization,
the experiments show that our dataset and tasks are learnable even for a smaller MLLM like Qwen3-VL-2B-Instruct.

\begin{figure*}[t]
  \includegraphics[width=\linewidth]{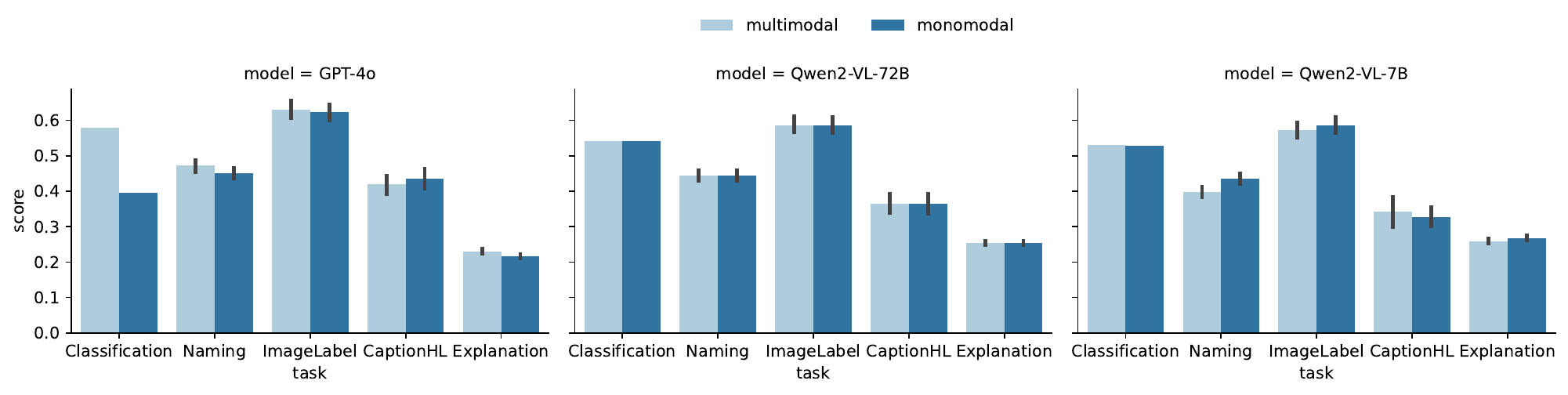}
  \caption{Model performance in multimodal versus monomodal experiments:
  average F1 score for Classification, sentence similarity score for Naming and ImageLabel, Jaccard index score for CaptionHL, ROUGE-1 for Explanation. Success rate is 100\% except for Qwen2-VL-7B-Instruct in the monomodal CaptionHL task (97\%).}
  \label{fig:multi-vs-mono}
\end{figure*}

\subsection{Ablation study}

To examine the models' processing of multimodal input in our tasks,
we design an ablation study that replaces image input with textual descriptions of the images, making the input data purely textual.
We use the \texttt{image\_description} data provided in the CapCon corpus, 
which are short, literal descriptions of the scene.
For example, the description for the image in Figure~\ref{fig:train-wreck} is as follows:
\enquote{A man and woman are in bed together under the covers. They are looking towards the bedroom door when a car has crashed into their home. They don't seem too upset about the situation.}

We rerun the six tasks on the same three models for prompt engineering:
GPT-4o and the two Qwen2-VL models.
For Classification, we use the prompt that results in the highest average F1 score for most models in prompt engineering: It asks the models to answer the question with \enquote{No or Yes} instead of \enquote{Yes or No}.

As shown in Figure~\ref{fig:multi-vs-mono}, there is not much difference between the models' performance in the multimodal and monomodal experiments, except for GPT-4o in the Classification task.
The comparison indicates that the multimodal input data is not adequately utilized by the models. Our error analysis provides more support for this.

\subsection{Error Analysis}

\begin{figure}[t]
  \includegraphics[width=\columnwidth]{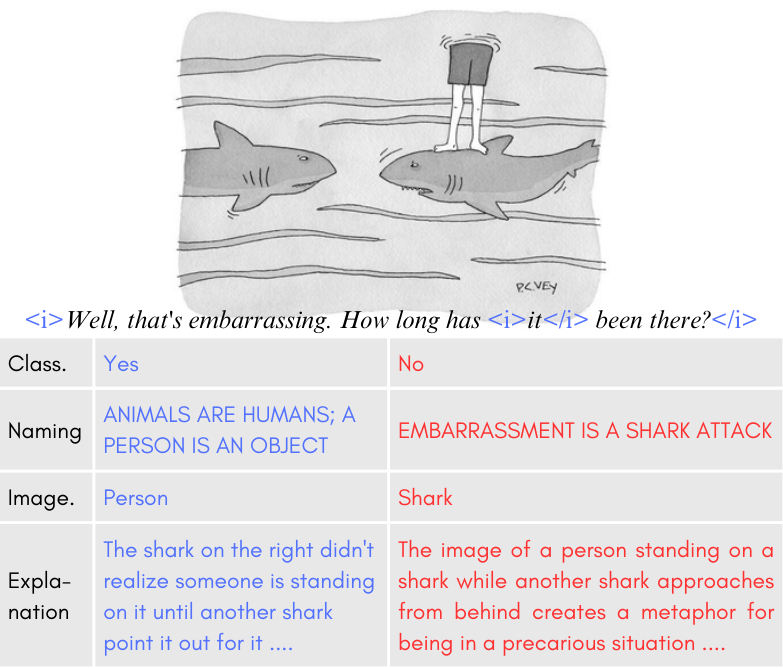}
  \caption{\textcolor{RoyalBlue}{Ground truth (left and in the caption)} versus \textcolor{red}{LLaVA-NeXT-110B predictions (right)} for the given image-caption pair. Explanations are shortened. For CaptionHL, the model outputs the entire caption as answer.}
  \label{fig:error-shark}
\end{figure}

\begin{figure}[t]
\centering
  \includegraphics[width=0.7\columnwidth]{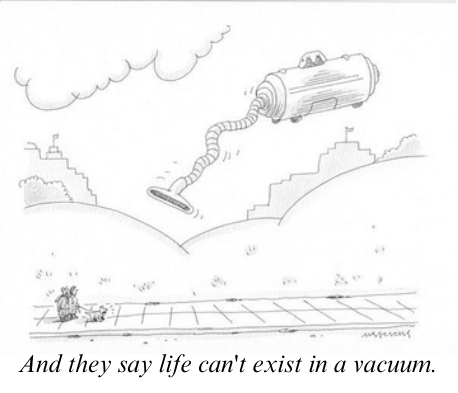}
  \caption{All models except GPT-4o predict \textsc{life is a vacuum} for this item in the Naming task.}
  \label{fig:error-vacuum}
\end{figure}

A primary reason for incorrect answers is the models' inability to integrate visual and textual information into a coherent story.
For example, the humor of the image-caption pair in Figure~\ref{fig:error-shark} is based on personification of the sharks:
They communicate in human language and can experience and express embarrassment.
Subsequently, the person in the image is compared metaphorically to an object, referred to as \lingform{it} in the caption.

LLaVA-NeXT-110B's answers indicate that the model assumes the caption is uttered by the person in the image, thus overlooking the \textsc{animals are humans} metaphor and the subsequent \textsc{a person is an object} metaphor.
This example also shows the importance of metaphor processing in humor understanding:
One cannot understand the humor of this image-caption pair without recognizing that the sharks are personified.

On the other hand, the models can usually identify common metaphorical or idiomatic expressions in the caption, although it does not guarantee adequate understanding of humorous multimodal metaphor use.
For example, the cartoon in Figure~\ref{fig:error-vacuum} depicts an alien spaceship as a vacuum cleaner.
The humor comes from the pun on \lingform{vacuum} in the caption:
It refers both to aliens physically existing in a vacuum cleaner, as depicted in the image, and life existing \lingform{in a vacuum} in an idiomatic sense.
The ground truth annotation of the conceptual metaphor is \textsc{alien spaceship is a vacuum cleaner}, because it is the base of the pun on \lingform{vacuum} in the caption.
The MLLMs acknowledge the metaphoricity of the expression \lingform{life can't exist in a vacuum}, but their prediction in the Naming task, \textsc{life is a vacuum}, indicates that they do not relate it to the vacuum cleaner spaceship in the image, thus failing to properly understand the humorous multimodal metaphor use. 

Our analysis indicates that the models' performance can potentially be improved when they are instructed more explicitly to combine image and caption information.
For example, one can use chain-of-thought prompting to instruct an MLLM to mimic the human annotation process: first process the image, and then integrate it with the caption, before proceeding to metaphor identification and understanding.

\section{Conclusion}

This study releases a dataset that provides expert annotation on humorous multimodal metaphor use.
Using the dataset, we benchmark popular and state-of-the-art MLLMs on their capabilities to identify and understand humorous multimodal metaphor use.
Our experiments show that current MLLMs struggle with processing humorous multimodal metaphor use, especially with regard to integrating visual and textual information.

\paragraph{Limitations and future directions.}
We hope this study will encourage more research on MLLMs' capabilities to process humorous multimodal metaphors.
As observed in Section~\ref{sec:model-human-compare-expt},
while the different annotation tasks inform each other for human experts,
MLLMs' performance may be further compromised when tackling multiple tasks at once.
We anticipate that reasoning models such as Qwen3-VL-32B-Thinking are more likely to benefit from this approach, but the evaluation and training of these models are beyond the scope of the current study.

Despite the expertise of our annotators,
the \textsc{Hummus} dataset does not account for potential cultural differences in humor and metaphor understanding.
Nevertheless, since the two annotators are from different cultures,
our dataset should feature humorous multimodal metaphors that are widely accepted in the field of metaphor research.

\section{Ethical Considerations}

We access the GPT-4 models through OpenAI API.
The open-source models (LLaVA-NeXT, Qwen2-VL, and Qwen3-VL) as well as the CapCon corpus are accessed through HuggingFace.
The CapCon corpus has CC-BY-4.0 license.
Our dataset and code
is freely accessible on GitHub.

Our dataset includes jokes that could be considered offensive, and certain jokes may be inappropriate for a younger audience.
These data remain in our dataset as they are a nonremovable part of multimodal humor and can be valuable for future research.

\bibliography{anthology,custom}
\bibliographystyle{acl_natbib}

\clearpage

\appendix

\section{Annotation Guidelines}
\label{sec:annotation-guidelines}

\begin{figure*}[t]
\centering
\begin{subfigure}{0.40\textwidth}
\centering
     \includegraphics[width=\textwidth]{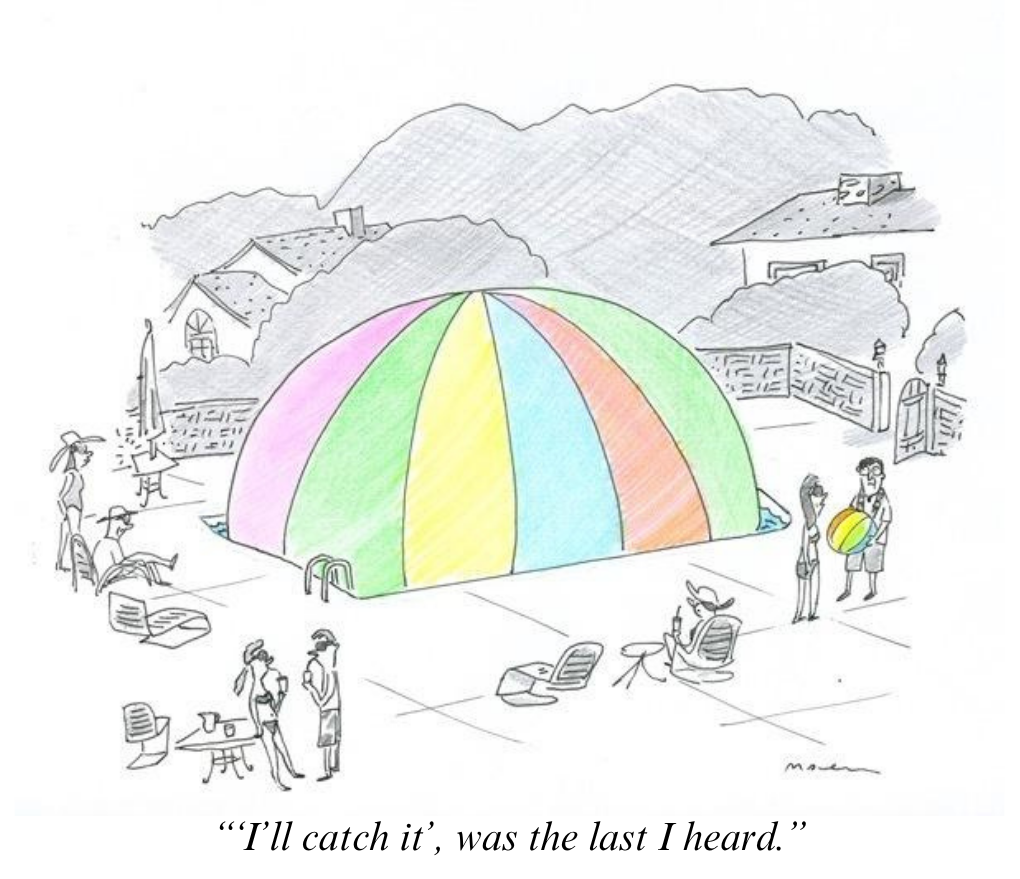}
    \caption{Non-metaphorical.}
    \label{fig:guidelines+nonmet}
\end{subfigure}
\begin{subfigure}{0.40\textwidth}
\centering
    \includegraphics[width=\textwidth]{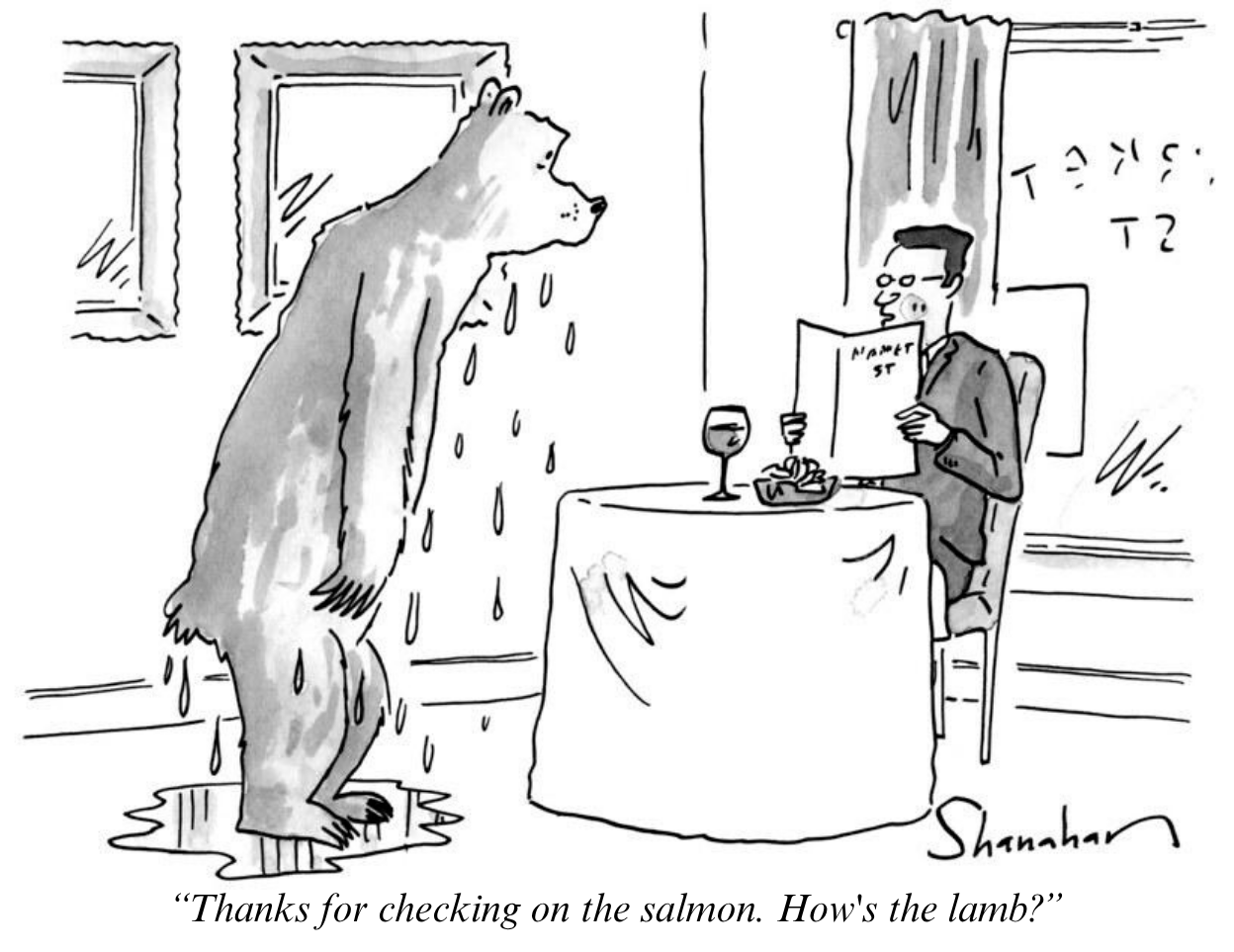}
    \caption{Metaphorical.}
    \label{fig:guidelines+met}
\end{subfigure}
\caption{Image-caption pair that (a) does not involve and (b) involves humorous metaphor use.}
\end{figure*}

\begin{figure*}[t]
\centering
\begin{subfigure}{0.32\textwidth}
\centering
     \includegraphics[width=\columnwidth]{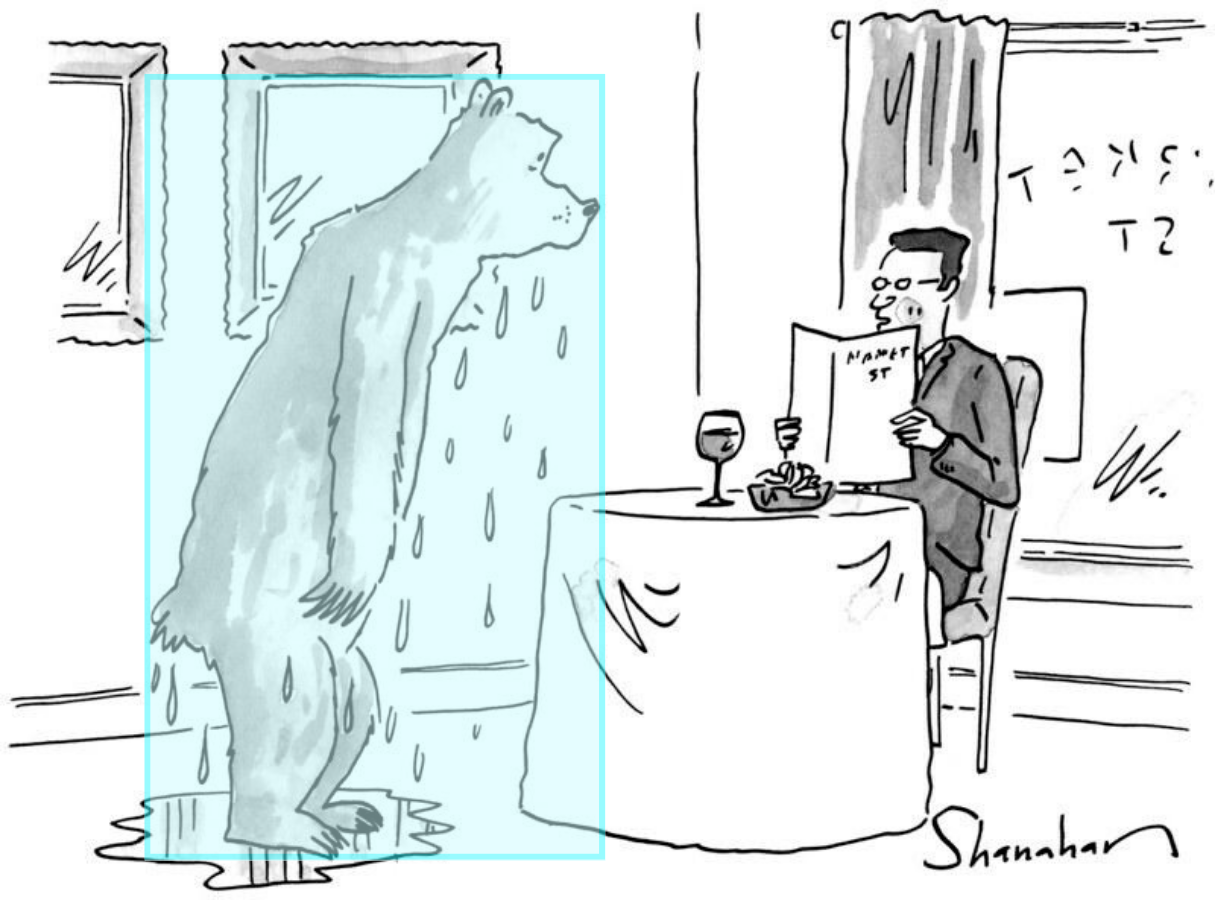}
    \caption{Correct.}
    \label{fig:bbox+correct}
\end{subfigure}
\hfill
\begin{subfigure}{0.32\textwidth}
\centering
    \includegraphics[width=\columnwidth]{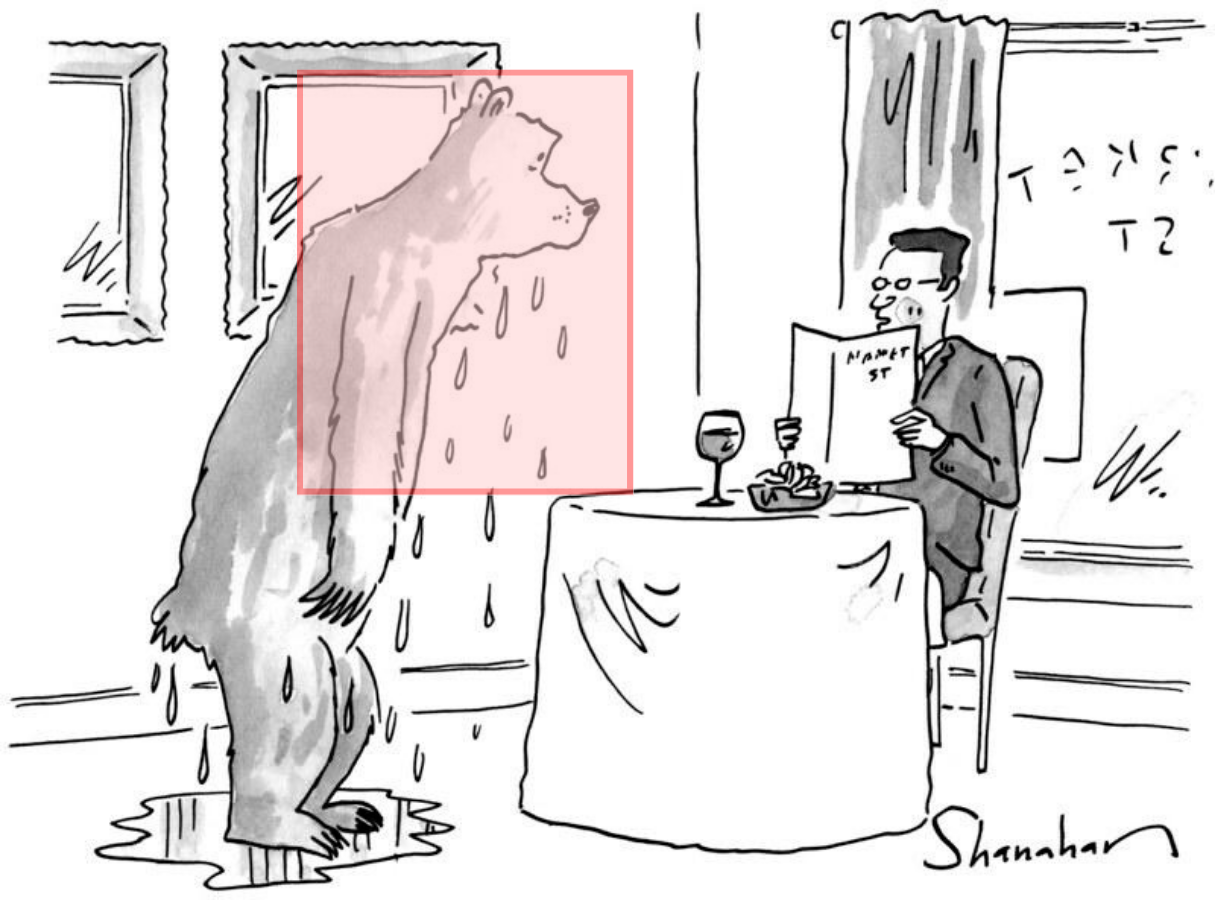}
    \caption{Too small.}
    \label{fig:bbox+small}
\end{subfigure}
\hfill
\begin{subfigure}{0.32\textwidth}
\centering
    \includegraphics[width=\columnwidth]{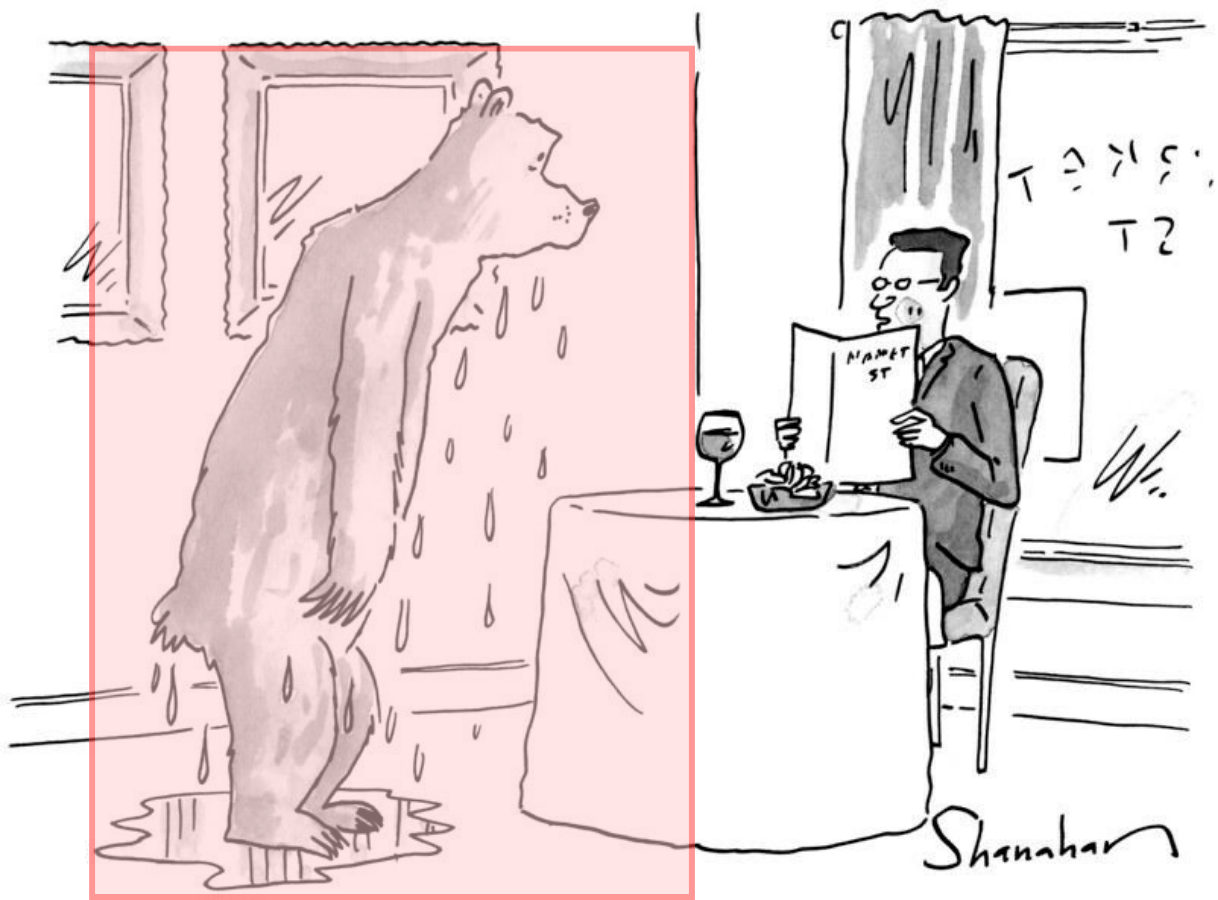}
    \caption{Too big.}
    \label{fig:bbox+big}
\end{subfigure}
\caption{Good and bad examples for bounding box annotation.}
\end{figure*}

\subsection{Objective}

You will annotate New Yorker cartoons and their captions in terms of humorous metaphor use.

\subsection{Labelling Steps}

\begin{enumerate}
    \item Look at the cartoon image alone (without referring to the caption). Find any incongruities in the image. You can write these down in the \texttt{incongruity} field, but this is optional.
    \item Copy and paste the caption to \texttt{text\_hl}.
    \item How does the caption resolve the incongruities in the image? If you can answer this question, then you understand the humor of the image-and-caption combination. You can write down your understanding in the \texttt{resolution} field, but this is optional.
    \begin{enumerate}
        \item If you understand the humor, continue with the annotation;
        \item Otherwise, select \enquote{Discard} for the \texttt{is\_met} field, assign \enquote{1} to \texttt{complete} and submit your annotation for this item.
    \end{enumerate}
    \item Based on your understanding of the humor, consider whether metaphor use (as detailed in \enquote{What counts as metaphor}) is involved in creating that humorous effect. Any kind of involvement counts. In essence, for each metaphor use you find in the image-and-caption combination, consider whether the humor will still be there when the metaphor use is removed.
    \begin{enumerate}
        \item If the answer is an absolute yes, assign \enquote{Yes} to \texttt{is\_met} and continue with the annotation;
        \item If the answer is an absolute no, assign \enquote{No} to \texttt{is\_met}, \enquote{1} to \texttt{complete} and submit your annotation;
        \item If you hesitate, assign \enquote{WIDLII} to \texttt{is\_met} and continue with the annotation.
    \end{enumerate}
    \item Specify the conceptual metaphor(s) in \enquote{X is Y} format. \enquote{X} should be the target domain; \enquote{Y} should be the source domain. Separate multiple conceptual metaphors with \enquote{; } (a semicolon followed by a whitespace).
    \begin{enumerate}
        \item To decide how general/specific the conceptual metaphor should be, try to relate the instantiation of the metaphor in the image-and-caption combination to other instantiations of the same metaphor (in any modality), and then find a term that can cover them all. 
        \item You can draw inspiration from this (non-exhaustive) \href{https://www.lang.osaka-u.ac.jp/~sugimoto/MasterMetaphorList/MetaphorHome.html}{list of conceptual metaphors}.
        \item Finishing the \texttt{image\_hl} and \texttt{text\_hl} annotation first (Step 6) could be helpful if you had difficulty pinpointing the two domains.
    \end{enumerate}
    \item Annotate the metaphor-related image areas (\texttt{image\_hl}) and text fragments (\texttt{text\_hl}):
    \begin{enumerate}
        \item To annotate metaphor-related image areas, try to assign each object/element of the image to either the target or the source domain of the conceptual metaphors you have annotated. Typically, the majority of the image should belong to the same domain (e.g., the target domain); that one object/element that belongs to the other domain (e.g., the source domain) is the metaphor-related image area. Draw a \texttt{image\_hl} bounding box for each such \enquote{odd} object/element (e.g., if the minority domain is CAT and there are 3 cats in the image, draw 3 bounding boxes, one box for each cat);
        \item Do the same for the caption. In the \texttt{text\_hl} field, where you already have the caption copied and pasted, surround each word/phrase/clause that belongs to the \enquote{odd} domain with a pair of <i></i> tag. If the entire caption really belongs to a single domain, surround the entire caption in a pair of <i></i> tag. Note that \texttt{image\_hl} and \texttt{text\_hl} do \emph{not} have to belong to the same domain.
    \end{enumerate}
\end{enumerate}

\subsection{Mandatory fields}

\begin{enumerate}
    \item Every item should have a \texttt{is\_met} value.
    \item If \texttt{is\_met} value is \enquote{Yes} or \enquote{WIDLII}, the following fields should also be filled: \texttt{image\_hl}, \texttt{text\_hl} (with <i>highlighted text</i>), and \texttt{conceptual\_metaphor}.
\end{enumerate}

\subsection{What counts as metaphor}

Following the Conceptual Metaphor Theory, we define metaphors as conceptual mappings between two difference domains. Therefore:

\begin{enumerate}
    \item Personification and zoomorphism are considered metaphor use. Personification is metaphor use with HUMAN as the source domain; zoomorphism is metaphor use with ANIMAL as the source domain.
    \item Metonymy within the same conceptual domain does not count as metaphor use.
    \item Idioms are not metaphors, unless the underlying cross-domain mapping is strictly required to make sense of the humor.
    \item Puns can indicate metaphor use. Whether they are metaphorical depends on whether the two meanings can be attributed to some sort of cross-domain mapping.
\end{enumerate}

\subsection{Non-metaphorical examples}

Example 1: Figure~\ref{fig:guidelines+nonmet}.
\textbf{Incongruity}: Giant beach ball.
\textbf{Resolution}: Whoever that was confident to catch the ball is now under it.
\textbf{Why not metaphorical}: It’s literally a giant ball; someone’s literally under it in the pool.

Example 2 and 3 are omitted due to space limit.

\subsection{Metaphorical examples}

Example 1: Figure~\ref{fig:guidelines+met}.
\textbf{Incongruity}: Bear in a restaurant.
\textbf{Resolution}: The bear is a waiter.
\textbf{Why metaphorical}: Restaurant waiters are human beings.
\textbf{Conceptual metaphor}: \textsc{animals are human workers}.
\textbf{Image annotation}: See Figure~\ref{fig:bbox+correct}.
\textbf{Caption annotation}: \enquote{Thanks for checking on the <i>salmon</i>. How's the lamb?}.

Example 2 and 3 are omitted due to space limit.

\subsection{Bounding box examples}

The bounding box should fit tightly around the object of interest---for example, the bear in Figure~\ref{fig:bbox+correct}.
Make sure the entire object is included; Figure~\ref{fig:bbox+small} is therefore incorrect.
But also avoid including excessive empty space around the object, such as Figure~\ref{fig:bbox+big}.

\section{Additional prompts and model outputs}
\label{sec:more-prompts}

\newcommand{\namingsample}[3]{#1 & \multirow{2}{*}{#2} \\ \emph{vs.} #3 & }

\begin{table}[t]
\centering
\begin{tabular}{m{5.3cm} >{\centering\arraybackslash}m{1.5cm}}
\toprule
 \textbf{Model Output \emph{vs.} Ground Truth} & \textbf{Score} \\
\midrule
\namingsample{Dogs are people}{0.83}{Animals are humans} \\
\midrule
\namingsample{Temperature is music}{0.73}{Temperature is pitch} \\
\midrule
\namingsample{Thoughts are bathwater}{0.63}{Psychotherapy is a bath} \\
\midrule
\namingsample{Parenting is egg incubation}{0.60}{Human baby is an egg} \\
\midrule
\namingsample{Humor is a drug}{0.55}{Preaching is a joke} \\
\midrule
\namingsample{Faith is a wedding}{0.45}{Alcohol is god}	\\
\midrule
\namingsample{Work is a tool}{0.35}{Psychotherapy is a bath} \\
\midrule
\namingsample{Slaying a dragon is a task}{0.26}{Modern man is knight} \\
\midrule
Ending a relationship is falling off a cliff \emph{vs.} Social media is physical world & 0.17 \\
\bottomrule
\end{tabular}
\caption{Sample model outputs in the Naming benchmark task  and their cosine similarity scores compared with ground truth.}
\label{tbl:scores-cm}
\end{table}

\begin{table}[t]
\centering
\begin{tabular}{llc}
\toprule
 \textbf{Model Output} & \textbf{Ground Truth} & \textbf{Score} \\
\midrule
Gun & Pistol & 0.86 \\
Teddy-bear & Toy bear & 0.75 \\
Meteor & Asteroid & 0.69 \\
Tree stump & Trunk & 0.62 \\
Ball & Egg & 0.57 \\
Hat & Cellphone & 0.47 \\
Alligator & Customer & 0.40 \\
Laptop & Asteroid & 0.32 \\
Beer & Human-shaped hole & 0.21 \\
Broom & Alien spaceship	& 0.10 \\
\bottomrule
\end{tabular}
\caption{Sample model outputs in the ImageLabel benchmark task and their cosine similarity scores compared with ground truth.}
\label{tbl:scores-imagelabel}
\end{table}

\begin{figure*}[t!]
  \includegraphics[width=\textwidth]{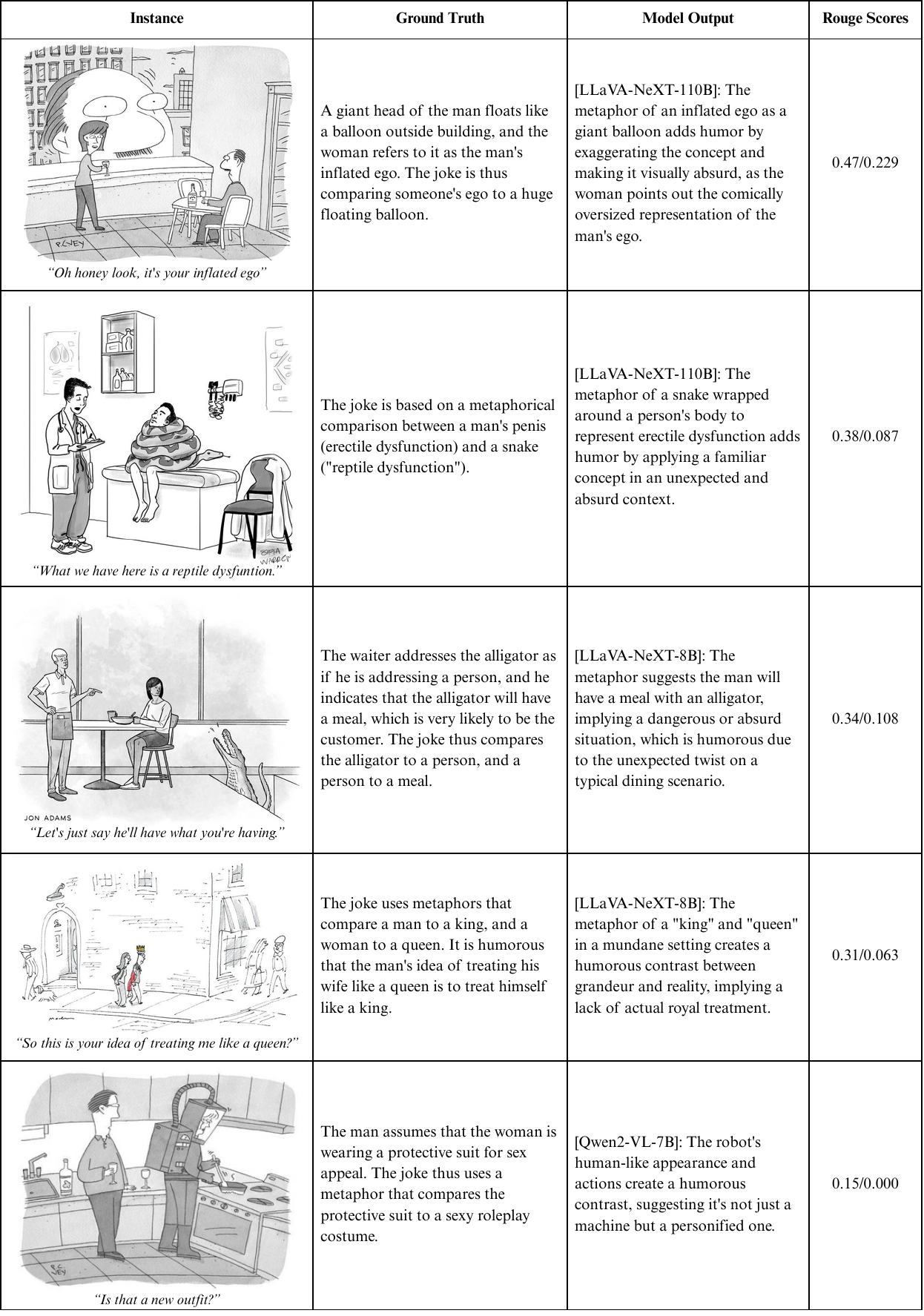}
  \caption{Sample model outputs in the Explanation benchmark task and their ROUGE-1/ROUGE-2 scores as compared to ground truth.}
  \label{fig:scores-explanation}
\end{figure*}

\begin{figure*}[t]
  \includegraphics[width=\textwidth]{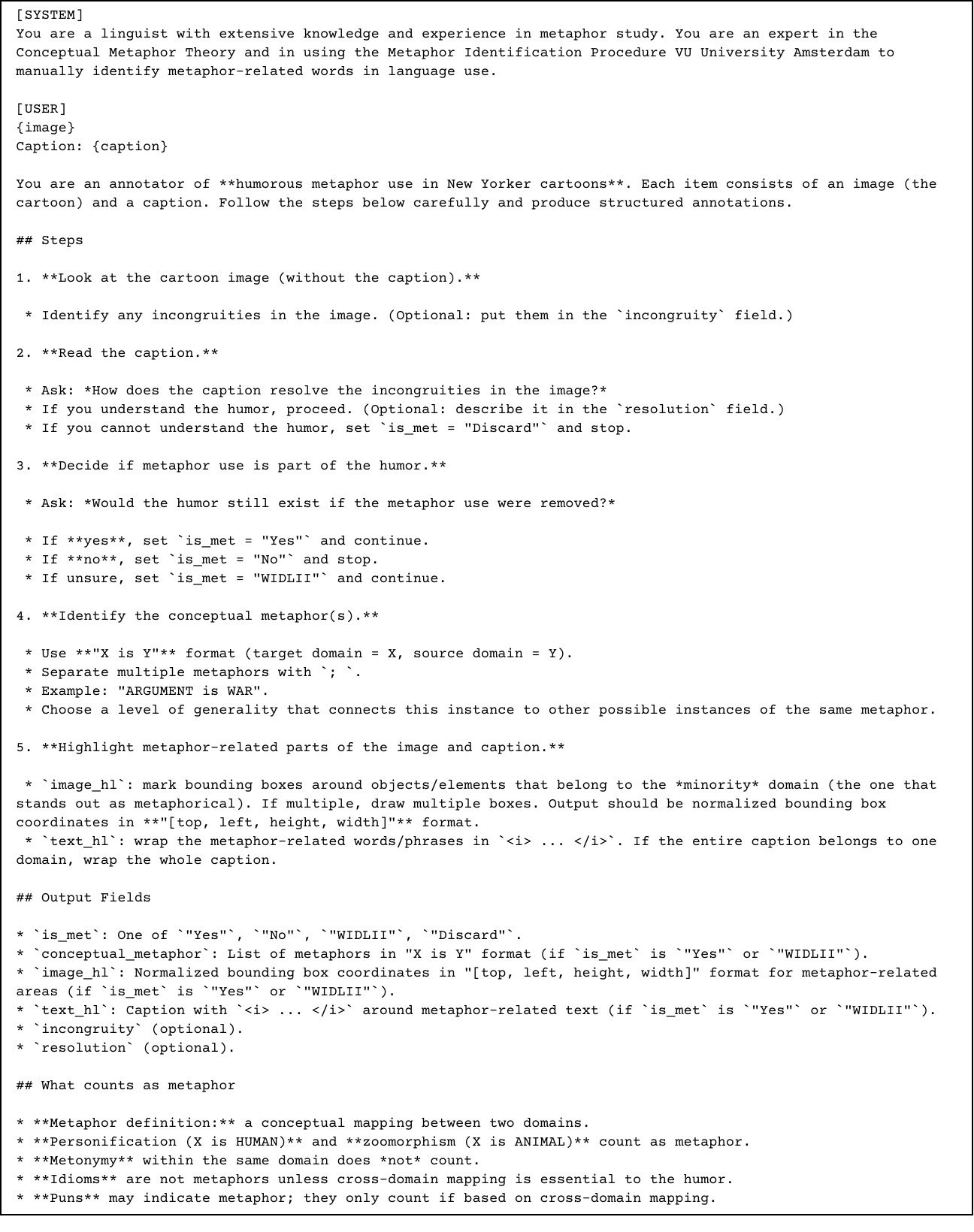}
  \caption{Prompt for comparing model and human performance.}
  \label{fig:prompt-aio03}
\end{figure*}

Table~\ref{tbl:scores-cm}, Table~\ref{tbl:scores-imagelabel}, and Figure~\ref{fig:scores-explanation} provide example model outputs in the ImageLabel, Naming, and Explanantion benchmark tasks respectively.

The prompt used in Section~\ref{sec:model-human-compare-expt} to compare model and human performance is provided in  Figure~\ref{fig:prompt-aio03}.

\end{document}